\documentclass{article}
\usepackage[final,nonatbib]{neurips}
\usepackage[utf8]{inputenc}
\usepackage{appendix}
\usepackage{enumitem}
\usepackage{hyperref}
\usepackage{bm}
\usepackage{mathtools}
\usepackage{multirow}
\usepackage{makecell}
\usepackage{float}
\usepackage{wrapfig}
\usepackage{xcolor}
\usepackage{amssymb}
\usepackage{amsmath}
\usepackage{booktabs}

\setlist[description]{leftmargin=0.75cm,labelindent=0.5cm}

\definecolor{diffusion}{HTML}{FF4500}
\definecolor{energy}{HTML}{8066AF}
\definecolor{synapse}{HTML}{FF6B00}
\definecolor{basin}{HTML}{6495ED}

\newcolumntype{M}[1]{>{$\displaystyle\qquad}p{#1}<{$}} %
\newcommand{\R}{\mathbb{R}}
\newcommand{\LL}{\mathcal{L}}
\newcommand{\vx}{\mathbf{x}}
\newcommand{\vv}{\mathbf{v}}

\newcommand{\vm}{\mathbf{m}}
\newcommand{\vw}{\mathbf{w}}

\newcommand{\veta}{\pmb{\eta}}

\newcommand{\vf}{\mathbf{F}}
\newcommand{\vff}{\mathbf{f}}
\newcommand{\vmu}{\pmb{\mu}}

\newcommand{\vtheta}{\boldsymbol{\theta}}
\newcommand{\mW}{\mathbf{W}}

\DeclareMathOperator{\sigmoid}{sigmoid}

\newcommand{\tinybrain}{\textsl{TinyBrain}} \begin{document}

\title{Memory in Plain Sight \\ \vspace{0.2cm} \large{Surveying the Uncanny Resemblances of Associative Memory and Diffusion Models}}
\author{
Benjamin Hoover \\
IBM Research\\ 
Georgia Tech\\
\texttt{benjamin.hoover@ibm.com}
\And
Hendrik Strobelt \\
MIT-IBM Watson AI Lab \\
IBM Research\\
\And
Dmitry Krotov \\
MIT-IBM Watson AI Lab \\
IBM Research\\
\And
Judy Hoffman \\
Georgia Tech\\
\And
Zsolt Kira \\
Georgia Tech\\
\And
Duen Horng Chau \\
Georgia Tech\\
}

\maketitle

\begin{abstract}
  The generative process of Diffusion Models (DMs) has recently set state-of-the-art on many AI generation benchmarks. 
  Though the generative process is traditionally understood as an ``iterative denoiser'', there is no universally accepted language to describe it.
  We introduce a novel perspective to describe DMs using the mathematical language of \textit{memory retrieval} from the field of energy-based Associative Memories (AMs), making efforts to keep our presentation approachable to newcomers to both of these fields. 
  Unifying these two fields provides insight that DMs can be seen as a particular kind of AM where Lyapunov stability guarantees are bypassed by intelligently engineering the dynamics (i.e., the noise and step size schedules) of the denoising process.
  Finally, we present a growing body of evidence that records DMs exhibiting empirical behavior we would expect from AMs, and conclude by discussing research opportunities that are revealed by understanding DMs as a form of energy-based memory.
\end{abstract}

\section{Introduction} \label{sec:introduction}

\begin{figure*}[t]
    \centering
    \includegraphics[width=\linewidth]{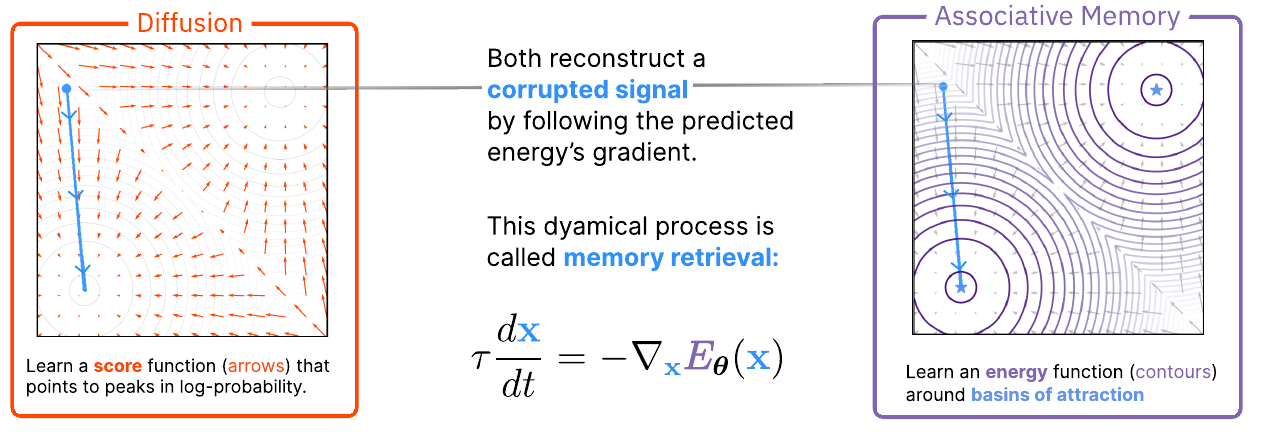}
    \caption{Comparing the emphases of \textcolor{diffusion}{Diffusion Models} and \textcolor{energy}{Associative Memories} tasked with learning the same energy (negative log-probability) landscape, represented with both contours and gradient arrows. \textcolor{diffusion}{Diffusion Models} (left) train a \textcolor{diffusion}{score function} (depicted as orange arrows) to model the gradient of the energy. The noisy starting signal (depicted as a blue circle) becomes less corrupted by following these gradients in the reverse denoising process. \textcolor{energy}{Associative Memories} (right) instead learn a smooth \textcolor{energy}{energy function}, depicted as \textcolor{energy}{contours}. The ``memory retrieval dynamics'' is the process by which a fixed point is retrieved by following the energy gradient from the initial signal.
    \textit{This process is mathematically equivalent to the objective of the reverse denoising process of Diffusion Models}. Memory retrieval dynamics always converge to \textcolor{basin}{fixed points} (there are two in each plot, one at the top right and lower left) where the energy is at a local minimum. This guarantee does not exist for Diffusion Models.}
    \label{fig:fig1}
\end{figure*}

Diffusion Models~\cite{sohl-dickstein2015Deep,song2019Generative,song2020Improved,ho2020Denoising} (DMs) have rapidly become the most perfomant class of generative models on images~\cite{dhariwal2021Diffusion,ramesh2022Hierarchical,saharia2022Photorealistic,rombach2022high,nichol2022GLIDE}. 
However, instead of decoding latent inputs into images using a single forward pass like both GANs~\cite{goodfellow2014generative} and VAEs~\cite{kingma2013AutoEncoding} do, DMs iteratively apply a denoising function that makes some ``latent'' representation of the image look more and more like the training distribution.
Training a DM consists of an unparameterized \textit{forward process} where a known amount of noise is repeatedly added to corrupt an image, and a \textit{reverse process} where the model learns to remove the noise added during the forward process. The computational power of the DM lies entirely in its reverse process, which is a sequence of denoising steps that aims to decrease the \textit{energy} (equivalent to increasing the log-probability, see \autoref{sec:diffusion-models}) of some noisy sample, effectively performing \textit{error correction}. 
The reverse process can generate novel images by ``error correcting'' pure noise.

In a seemingly unrelated field, Associative Memories (AMs) are formalized as dynamical systems that are concerned with the storage and retrieval of data (a.k.a. ``information'' or ``signals'')~\cite{hopfield1982Neural,hopfield1984Neurons}. 
Like other Energy Based Models~\cite{lecunTutorial}, AMs are described by an explicit and learnable energy function that places corrupted signals at high energy and uncorrupted or ``real-looking'' signals at low energy. The ensuing dynamics minimize the energy of an initial signal, a process that performs \textit{error correction} on that signal. A signal is said to be ``stored'' (memorized) if it lives at a local minimum of the energy landscape. These local minima are called \textit{memories}~\cite{krotov2021large}.

How do AMs retrieve memories? We can think of the energy landscape as a hilly terrain, where a ``ball'' is placed at some initial location (i.e., our corrupted signal acting as a ``query'').
The ball rolls down the hill (the query evolves, becoming less corrupted), eventually settling at a local minimum (a memory) according to the dynamical equation and diagram in \autoref{fig:fig1}. 
This process is formally called \textit{memory retrieval}. \textbf{Memory retrieval looks eerily like the reverse process of DMs}.
Both approaches have the same goal of error correction:

\vspace{0.2cm}
\begin{quote}
\textbf{Goal of both DMs and AMs}: \textit{Given a corrupted representation of some data, recreate the original uncorrupted data by descending some energy.}
\end{quote}
\vspace{0.2cm}

\noindent Yet, though DMs have been related to Markovian VAEs~\cite{mcallester2023Mathematics,ho2020Denoising,sohl-dickstein2015Deep}, Normalizing Flows~\cite{grathwohl2018FFJORD}, Neural ODEs~\cite{song2020ScoreBased}, and generic Energy Based Models \cite{song2019Generative,lippe2021Tutorial}, an explicit connection to AMs has not been fully explored in part due to the lack of awareness about AMs.
Such a connection would contribute to a growing body of literature that seeks to use modern AI techniques to unravel the mysteries of memory and cognition~\cite{takagi2023high,whittington2022relating,fu2023Pattern,ozcelik2023BrainDiffuser,wang2022Incorporating,majumdar2023Where,kozachkov2023building}.

\subsection{Related Surveys} \label{sec:related-surveys}

The popularity of DMs has resulted in surveys that focus on the methods and diverse applications of DMs~\cite{yang2022Diffusion,song2019Generative,ma2022Pedagogical} alongside tutorial-style guides that seek to gently explain them~\cite{song2019Generative,luo2022Understanding,kreisDenoising}. 
In all cases, these surveys/guides make no connection to AMs. For an exhaustive reference on DM literature, \cite{thorton2023What} has collected a (still growing) list of ${>}600$ diffusion papers.
Other surveys cover AMs and their applications~\cite{sulehria2007hopfield,hanlon1966ContentAddressable,prasad2012Study,gritsenko2017Neural,s.a2018Survey}, while others acknowledge high-level similarities between recurrent networks and AMs~\cite{poggio2021Associative,krotov2023new}. We are aware of a concurrent preprint by~\cite{ambrogioni2023search} that is the closest in spirit to this work and whose valuable contributions we discuss in \autoref{sec:math-connection}. Our survey provides a more thorough focus on AMs while emphasizing an approachable introduction to DMs without relying on stochasticity during memory retrieval~\cite{song2020ScoreBased}.

\subsection{Our Contributions} \label{sec:contributions}
This survey examines the striking similarity between Diffusion Models and Associative Memories,
straddling a gap between traditional and modern AI research that is rarely bridged. This survey:

\begin{enumerate}[leftmargin=*]
    \itemsep0em
    \item \textbf{Provides an approachable overview} of both AMs and DMs from the perspective of dynamical systems, energy, and Ordinary Differential Equations (ODEs) (\autoref{sec:diffusion-models} and \autoref{sec:associative-memories}). 
    We raise awareness about AMs by leaning into the popularity of DMs while simultaneously distilling the traditionally complex presentation of DMs using more intuitive descriptions of memory retrieval.
    We isolate the differences between DMs and AMs (e.g., AM architectures satisfy Lyapunov stability criteria that DMs do not) and discuss evidence that the differences are mitigated through the design and usage of DMs (\autoref{sec:similarities}).
    \item \textbf{Highlights future research directions} at the intersection of DMs and AMs, made possible by characterizing their resemblance (\autoref{sec:conclusions}).
    We also identify similarities that AMs have to other modern architectures (e.g., Transformers~\cite{vaswani2017attention,ramsauer2022Hopfield,hoover2023Energy}). These similarities highlight mounting evidence that the field of AI is converging to models that strongly resemble AMs, escalating the urgency to understand Associative Memories as an eminent paradigm for computation in AI.
\end{enumerate} %
\section{Mathematical Notations}
\label{sec:math-notations}

In this survey we deviate from notations typically used for DMs and AMs. To minimize visual clutter, we prefer tensor notation over Einstein notation, representing scalars in non-bolded font (e.g., energies $E$ or time $t$), and tensors in bold (e.g., state vector $\vx$ or weight matrices $\mW$). These distinctions also apply to scalar- and tensor-valued functions (e.g., energy scalar $E(\cdot)$ vs. activation vector $\hat{\vx}(\cdot)$). A collection of learnable parameters is expressed through the generic variable $\vtheta$. 
Gradients are expressed using ``nabla'' notation of a scalar valued function, where $\nabla_{\vx} (\cdot)$ will be of the same shape as $\vx$. The transpose is represented using $\vx^\intercal$ notation, whereas $\vx^T$ represents the vector $\vx$ occurring at time $T$. 
\section{Diffusion Models} \label{sec:diffusion-models}

As generative models, Diffusion Models (DMs) seek to synthesize new data by sampling from some learned approximation $p_{\vtheta}$ of the data's probability density function (p.d.f.) $p_\text{data}$. 
However, rather than learn the likelihood $p_{\vtheta}$ itself, DMs use their parameters to approximate the gradient of the log-likelihood a.k.a. the \textbf{score function} $\vf_{\vtheta}(\vx) \triangleq \nabla_\vx \log p_{\vtheta} (\vx)$; thus, DMs are considered a class of \textit{score-based models}~\cite{song2019Generative,song2020ScoreBased,zimmermann2021ScoreBased,salimans2021Should}.

Mathematically, any p.d.f. can be expressed in terms of an \textbf{energy function} $E_{\vtheta} (\vx)$ by the Boltzmann Distribution

\begin{equation} \label{eq:boltzmann-dist}
   p_{\vtheta}(\vx) = \frac{e^{-E_{\vtheta} (\vx)}} {Z_{\vtheta}},
\end{equation}

\noindent where $Z_{\vtheta}$ is the partition function (i.e., the normalizing constant enforcing that $\int p_{\vtheta}(\vx) d\vx = 1$). Energy $E_{\vtheta}$ is always proportional to the log likelihood ($\log p_{\vtheta}$) and the score function is defined as the negative gradient of the energy:

\begin{equation} \label{eq:score-function}
    \vf_{\vtheta}(x) \triangleq \nabla_\vx \log p_{\vtheta} (\vx) = -\nabla_\vx E_{\vtheta}(\vx).
\end{equation}

\autoref{fig:fig1} depicts the score function as vectors pointing to peaks in the log probability. 
We often think of the score function as predicting the noise we need to remove from a data point $\vx^t$ at some time $t$, where \textit{adding} the estimated score $\vf_{\vtheta}(\vx^t)$ in \autoref{eq:score-dynamics} is the same as \textit{removing} the predicted noise. Thus, given a neural network trained to predict the score, the process of generating data using score-based models can be construed (in discrete time) as an iterative procedure that adds the predicted score (subtracts the predicted energy gradient) for some fixed number of steps $T$ on pure noise $x^0$. The final state $\vx^T$ is declared to be a local peak (minima) of the log-likelihood (energy) and should now look like a sample drawn from the original distribution $p_\text{data}$. This process is described in \autoref{eq:score-dynamics}, where $\alpha \in \mathbb{R}$ is a step size in the direction $F_{\vtheta}$:

\begin{equation} \label{eq:score-dynamics}
    \vx^{t+1} = \vx^t + \alpha \vf_{\vtheta}(\vx^t), \;\;\;\;\;\; t = 0, \ldots ,T-1.
\end{equation}

To be more precise, DMs use a score function $\vf_{\vtheta}(\vx;t)$ and step size $\alpha(t)$ (a.k.a. the ``scheduler'') that are both conditioned on time $t$~\cite{song2019Generative}. Also, note that \autoref{eq:score-dynamics} uses the convention that \emph{time progresses forward when reconstructing the data}. However, the literature around DMs describes \emph{time in the reconstruction process as going backwards}, denoting $x^T$ to refer to the sample drawn from pure noise and $x^0$ to refer to the final reconstructed sample drawn from $p_{\vtheta}$~\cite{sohl-dickstein2015Deep,song2019Generative,ho2020Denoising,mcallester2023Mathematics,song2020ScoreBased}. \autoref{eq:score-dynamics-reverse} rewrites \autoref{eq:score-dynamics} using the variable $s \triangleq T-t$ to represent the reverse-time convention used in most DM papers:

\begin{equation} \label{eq:score-dynamics-reverse}
    \vx^{s-1} = \vx^{s} + \alpha(s) \vf_{\vtheta}(\vx^s; s), \;\;\;\;\;\; s = T,\ldots,1.
\end{equation}

Using score-based models is then conceptually very simple: they seek to maximize (minimize) the log-likelihood (energy) of an initial sample by following the score $\vf_{\vtheta}$. 
However, DMs require several tricks to train, with the most popular being the technique of \textit{denoising score-matching}~\cite{song2020Sliced,song2019Generative,hyvarinen2005Estimation,raphan2011Least,vincent2011Connection,ho2020Denoising}. 
To train with denoising score-matching, samples $\vx$ from our original data distribution $p$ are repeatedly perturbed with small amounts of (time-dependent) noise $\veta(s)$. We then train our score-based model $\vf_{\vtheta}(\vx^{s+1};s)$ to remove the added noise.
If the noise is small enough everywhere, we can guarantee that our optimal score function (parameterized by optimal weights $\vtheta^*$) approximates the score of the true data distribution: i.e., $\vf_{{\vtheta}^*}(\vx) \approx \nabla_\vx \log p_\text{data}(\vx)$ and $\vf_{{\vtheta}^*}(\vx^{s+1}; s) \approx -\veta(s)$.

\subsection{Diffusion Models are Continuous Neural ODEs}
The original DMs~\cite{sohl-dickstein2015Deep,song2019Generative} relied on a fixed number of discrete and stochastic steps in the forward and reverse processes. This changed when \cite{song2020ScoreBased} introduced a \textit{probability flow Ordinary Differential Equation} (PF-ODE) formulation for DMs that formulated DMs in continuous time along deterministic trajectories.

Consider the standard form of a generic ODE under constrained time $s$

\begin{equation} \label{eq:basic-ode}
\frac{d\vx}{ds} = \vmu(\vx;s), \;\;\;\;\;\; s \in [T,0],
\end{equation}

\noindent where $\vmu(\vx;s)$ is an arbitrary \textit{drift function} representing some deterministic change in position of particle $\vx^s$ at time $s$.
DMs need to further corrupt the input $\vx$, so we add to \autoref{eq:basic-ode} an infinitesimal amount of noise $\frac{d\vw}{ds}$ scaled by some real-valued and time-dependent \textit{diffusion coefficient} $\sigma_s$. The forward process of PF-ODEs is thus an Itô Stochastic Differential Equation (SDE)~\cite{song2020ScoreBased}:

\begin{equation} \label{eq:basic-ode-corruption}
\frac{d\vx}{ds} = \vmu(\vx;s) + \sigma_s \frac{d\vw}{ds}.
\end{equation}

\noindent \cite{karras2022Elucidating} argues that the equation above can be further simplified without any loss in performance by assuming constant drift function $\vmu(\vx;s) = 0$, a convention adapted by \cite{song2023Consistency} 
to set SOTA one-step generation with DMs. This convention simplifies \autoref{eq:basic-ode-corruption} to make the forward process depend only on the diffusion coeffcient and the infinitesimal random noise, as in 

\begin{equation} \label{eq:simple-ode-forward}
\frac{d\vx}{ds} = \sigma_s \frac{d\vw}{ds}.
\end{equation}

\noindent The reverse process now depends only on the noise scale $\sigma_s$ and the score $\vf_{\vtheta}$~\cite{song2020ScoreBased,song2023Consistency}:

\begin{equation} \label{eq:simple-ode-reverse}
\frac{d\vx}{ds} = -\frac{1}{2} \sigma^2_s \vf_{\vtheta}(\vx; s).
\end{equation}

\noindent We have written the strange ``reverse time'' convention of DMs using time variable $s \triangleq T-t$. \autoref{eq:continuous-diffusion-energy} rewrites \autoref{eq:simple-ode-reverse} using forward time $t$, collecting the noise scale into a real-valued time-variable $\tau_t \triangleq \frac{2}{\sigma^2_t}$ to control the rate of change:

\begin{equation} \label{eq:continuous-diffusion-energy}
\tau(t)\frac{d\vx}{dt} = \vf_{\vtheta}(\vx; t)\,, \;\;\;\;\;\; t \in [0,T].
\end{equation}

PF-ODEs generalize previous theories of DMs by removing the mathematical dependency on discrete time and stochasticity in the reverse process. At the same time, the continuous dynamics of PF-ODEs exposes a strong mathematical connection to Associative Memories and deterministic memory retrieval that was difficult to see before.

\section{Associative Memories} 
\label{sec:associative-memories}

\begin{wrapfigure}[26]{r}{0.6\textwidth}
    \centering
    \includegraphics[width=\linewidth]{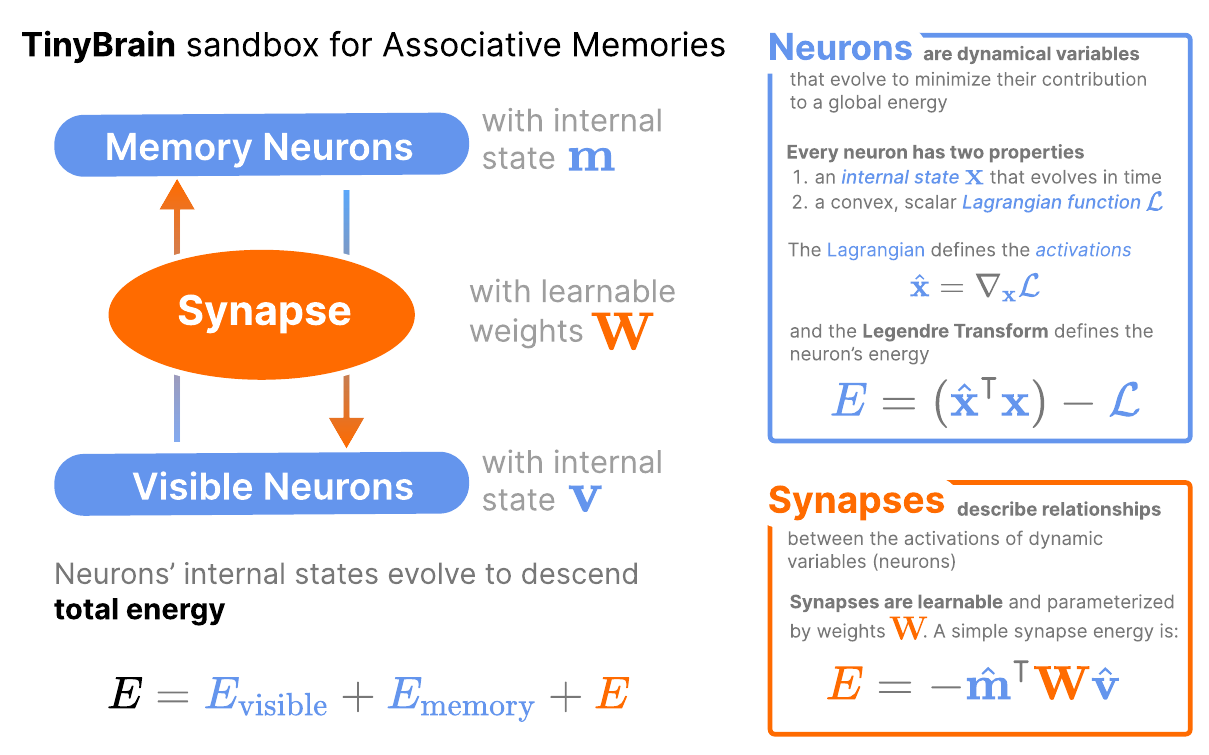}
    \caption{The {\tinybrain} sandbox for understanding Associative Memories is a fully connected bipartite graph (structurally similar to the Restricted Boltzmann Machine~\protect\cite{hinton2002Training}). \textcolor{basin}{Visible neurons} and \textcolor{basin}{memory neurons} have states $\vv$ and $\vm$ respectively that evolve in time; these states have corresponding activations $\hat{\vv}$ and $\hat{\vm}$. 
    The energy of the \textcolor{synapse}{synapse} is minimized when the memory activations $\hat{\vm}$ perfectly align with the visible activations $\hat{\vv}$ according to the learned parameters $\mW$. 
    }
    \label{fig:tinybrain}
\end{wrapfigure}

Associative Memory (AM) is a theory for the computational operation of brains that originated in the field of psychology in the late 19th century \cite{james1890principles}.
We are all familiar with this phenomenon. Consider walking into a kitchen and smelling a freshly baked pie. The smell could elicit strong nostalgia of holidays spent with extended family, which surfaces memories of the names and faces of everyone gathered. The smell (query) retrieved a set of feelings, names, and faces (values) from your brain (the associative memory). 
Because the query itself is of the same form as the data that was stored, AMs are considered a form of \textit{Content Addressable Memory} (CAM)~\cite{s.a2018Survey,sharma2022Content}.

AMs are energy functions $E_{\vtheta} \in \R$ that define dynamical systems capable of storing and retrieving data $\vx$ inside learned parameters $\vtheta$. 
Given any input signal $\vx^0$ at time $t=0$, we want to minimize the energy to a fixed point that represents a memory learned from the original data,

\begin{equation}
\label{eq:continuous-am-dynamics}
    \tau \frac{d\vx}{dt} = -\nabla_{\hat{\vx}} E_{\vtheta}(\hat{\vx})\,, \;\;\; t > 0.
\end{equation}

\noindent What is this new variable $\hat{\vx}$ doing in the energy function? $\hat{\vx}(\vx)$ is a function of hidden state $\vx$ and represents the \textit{conjugate variable} of $\vx$ as defined through the Legendre transform of a convex Lagrangian~\cite{krotov2021large, krotov2021Hierarchical}.
We explain this in more detail in \autoref{sec:am-constraints}, but for now we can think of $\hat{\vx}$ as an \textit{activation function} of $\vx$ that ensures the energy function describes a fixed-point attractor system.
\autoref{eq:continuous-am-dynamics} can of course be discretized and treated as a neural network that is recurrent through time:

\begin{equation}
\label{eq:discrete-am-dynamics}
    \vx^{t+1} = \vx^t - \frac{dt}{\tau} \nabla_{\hat{\vx}} E_{\vtheta}(\hat{\vx}^t).
\end{equation}

\noindent The energy is constructed to \textit{guarantee} convergence to a fixed point $\vx^\star$ at some $t=T$, when

\begin{align*}
\label{eq:fixed-point-continuous}
    \frac{d\vx^\star}{dt} = 0\,, \hspace{0.4cm} {\color{gray}\mathrm{and}} \hspace{0.4cm}  
    \vx^\star = \vx^\star - \frac{dt}{\tau} \nabla_{\hat{\vx}} E_{\vtheta}(\hat{\vx}^\star),\;\;\; t \geq T\,.
\end{align*}

\subsection{A Sandbox for Building Associative Memories}
\label{sec:am-constraints}

AMs are constrained to use only neural architectures that themselves have a \textit{Lyapunov function}: that is, these architectures must compute a scalar energy and their temporal behavior must lead to fixed point attractors. However, not all scalar energies are Lyapunov functions: in this section we discuss the architectural constraints needed to ensure Lyapunov stability, following the abstractions of~\cite{hoover2022Universal}.

AMs were originally inspired by neurons and synapses within the brain~\cite{hopfield1982Neural}. As such, it is useful to consider a simple model we call the {\tinybrain} model\footnote{Our use of the term ``brain'' does not claim resemblance to real brains or brain-like behavior; instead, the term encapsulates a useful thought experiment around the terms ``neurons'' and ``synapses''.} as depicted in \autoref{fig:tinybrain}.
The {\tinybrain} model is a bipartite graph consisting of two dynamic variables (``\textit{neuron layers}'') connected to each other via a single weight matrix $\mW$ (a ``\textit{synapse}'' describing the relationship between variables). Each neuron layer can be described by an \textit{internal state} $\vx \in \R^N$ (which one can think of as the \textit{membrane voltage} for each of the $N$ neurons in that layer), and their \textit{axonal state} $\hat{\vx} \in \R^{N}$ (which is analogous to the \textit{firing rate} of each neuron). 
We call the axonal state the \textit{activations} and they are uniquely constrained to be the gradient of a scalar, convex Lagrangian function $\LL: \R^N \mapsto \R$ defined on that layer; that is, $\hat{\vx}= \nabla_\vx \LL$.

The \textit{Legendre Transform} of the Lagrangian $\mathcal{L}$ defines the energy $E^\mathrm{layer}(\hat{\vx}) \in \R$ of a neuron layer characterized by hidden state $\vx$, shown in \autoref{eq:neuron-energy}~\cite{krotov2021large,krotov2021Hierarchical,hoover2022Universal}.
\textit{If all layers have a Lagrangian, a synaptic energy can be easily defined such that the entire system is a global Lyapunov function}.

\begin{equation}
\label{eq:neuron-energy}
    E^\mathrm{layer} = \hat{\vx}^\intercal \vx - \LL(\vx)
\end{equation}

\noindent For historical reasons~\cite{krotov2017dense}, we call one neuron layer in {\tinybrain} the ``visible neurons'' (fully described by the internal state $\vv$ and Lagrangian $\LL_v$) and the other the ``memory neurons'' (fully described by the internal state $\vm$ and Lagrangian $\LL_m$). These layers are connected via a synaptic weight matrix $\mW \in \R^{N_m \times N_v}$, creating a synaptic energy $E^\mathrm{synapse}(\hat{\vv}, \hat{\vm}; \mW) \in \R$ defined as 

\begin{equation}
\label{eq:synapse-energy}
    E^\mathrm{synapse} = -\hat{\vm}^\intercal \mW \hat{\vv} \, .
\end{equation}

We now have defined all the \textit{component energies} necessary to write the total energy $E^\mathrm{system}$ for {\tinybrain} as in \autoref{eq:tiny-brain-energy}. Note that \autoref{eq:tiny-brain-energy} is a simple summation of the energies of each component of the {\tinybrain}: two layer energies and one synaptic energy. 

\begin{equation}
\label{eq:tiny-brain-energy}
    E^\mathrm{system} = E^\mathrm{layer}_v + E^\mathrm{layer}_m + E^\mathrm{synapse}.
\end{equation}

\noindent The hidden states of our neurons $\vv$ and $\vm$ evolve in time according to the general update rule

\begin{align}
\label{eq:tiny-brain-update-rule}
   \begin{dcases}
        \tau_v \frac{d\vv}{dt} &= -\nabla_{\hat{\vv}} E^\mathrm{system} \vspace{0.1cm} \\
        \tau_m \frac{d\vm}{dt} &= -\nabla_{\hat{\vm}} E^\mathrm{system}\,.
    \end{dcases} 
\end{align}

\noindent \autoref{eq:tiny-brain-update-rule} reduces to the manually derived update equations for the visible and memory layers presented in \autoref{eq:tiny-brain-manual-update-rule}~\cite{krotov2021large},

\begin{equation}
\label{eq:tiny-brain-manual-update-rule}
   \begin{dcases}
        \tau_v \frac{d\vv}{dt} &= \mW^\intercal \hat{\vm} - \vv \vspace{0.1cm} \\
        \tau_m \frac{d\vm}{dt} &= \mW \hat{\vv} - \vm\,,
    \end{dcases}
\end{equation}

\noindent where $\mW^\intercal \hat{\vm}$ and $\mW \hat{\vv}$ represent the \textit{input currents} to visible neurons and memory neurons respectively, and $-\vv$ and $-\vm$ represent exponential decay (this implies that the internal states $\vv$ and $\vm$ will exponentially decay to $\mathbf{0}$ in the absence of input current). 

Krotov and Hopfield \cite{krotov2021large} identified that this toy model is mathematically equivalent to the original Hopfield Network~\cite{hopfield1982Neural,hopfield1984Neurons} under certain choices of the activation function $\hat{\vm}$; hence, it is through the lens of this abstraction that we explore the long history of AMs.

\subsection{Hopfield Nets are Energy-Based Associative Memory}
\label{sec:memory-capacity}

John Hopfield formalized dynamic Associative Memories in the 1980s using an energy-based model that became famously known as the \textit{Hopfield Network} (HN)~\cite{hopfield1982Neural,hopfield1984Neurons}.
The original HN, a.k.a. the \textit{Classical Hopfield Network} (CHN), was a two-layer {\tinybrain} with $N_v$ visible neurons and $N_m$ memory neurons connected via a synaptic weight matrix $\mW \in \R^{N_m \times N_v}$. Visible neurons $\vv$ had corresponding activations $\hat{v}_i = \sigmoid(v_i)$ (i.e., elementwise operation for $i \in \{1, \ldots, N_v \}$) and memory neurons $\vm$ had a linear activation function $\hat{m}_\mu = m_\mu$, for $\mu \in \{1, \ldots, N_m \}$. 

The CHN unfortunately suffered from a small memory capacity that scaled linearly according the number of visible neurons $N_v$; specifically, the maximum storage capacity of the CHN was discovered to be ${\sim}0.14N_v$ by~\cite{abu-mostafa1985Information,hopfield1982Neural,amit1985SpinglassModels}. Consider the problem of building an associative memory on the $60$k images in MNIST,
where each image can be represented as a binary vector with $N_v = 784$ features. Assuming random patterns, one could store a maximum of $0.14(784) \approx 110$ images using the CHN, \textbf{no matter how many memories $N_m$ you add to the synaptic matrix $\mW$}.
The capacity problem of the CHN severely limits its applications, though the design of the CHN inspired similarly famous architectures like the Boltzmann Machine~\cite{hinton1984boltzmann} and the Restricted Boltzmann Machine~\cite{hinton2002Training} and made a large impact in the statistical physics and neuroscience communities.

A breakthrough in the capacity of HNs was proposed by Krotov \& Hopfield over 30 years later~\cite{krotov2017dense}. Their network, called the Dense Associative Memory (DAM), enabled the energy dynamics of the CHN to store a super-linear number of memories. The core idea was to use a rapidly growing non-linear activation function $\hat{\vm}$ on the memory neurons. For instance, choosing $\hat{\vm}$ to be higher orders of (optionally rectified) polynomials allowed much greater memory storage capacity than the CHN. 
The intuition is that the ``spikier'' the activation function $\hat{\vm}$, the narrower the energy basin around $\vm$, and the more memories the network can store and retrieve.  
Using exponential functions $\hat{m}_\mu = \exp(m_\mu)$ for $\mu \in \{1, \ldots, N_m \}$ even leads to exponential storage capacity \cite{demircigil2017Model}. 
Recall that a CHN could reliably store a maximum of $110$ MNIST images. Using the exponential DAM, one can increase the number of stored memories up to $N_m \sim \exp(784/2)$ (assuming no correlations) \textit{and still reliably retrieve each individual memory}. This marked difference has led to DAMs being branded as the ``Modern Hopfield Network'' (MHN)~\cite{ramsauer2022Hopfield} and opened the door for a new frontier in AMs \cite{krotov2023new}.

\subsection{Adding Hierarchy to Associative Memories}
\label{sec:hierarchical-ams}

The CHN and DAM have another fundamental limitation: like the {\tinybrain}, both only have a single weight matrix $\mW$ and cannot learn hierarchical abstractions that simplify the task of memorizing complex signals.
However, \autoref{eq:neuron-energy} makes it easy to constrain the energy of an AM with arbitrary numbers of dynamic variables. Why can't these systems include more interacting layers than the {\tinybrain} in \autoref{sec:am-constraints}? This is the realization of \cite{krotov2021Hierarchical,hoover2022Universal} who generalize the abstraction of AMs in \autoref{sec:am-constraints} to invent the Hierarchical Associative Memory (HAM): a single energy composed of arbitrary numbers of neuron layers and complex synaptic operations that can resemble the convolutional, pooling, or even attention operations in modern architectures~\cite{hoover2023Energy}. 

The HAM has given AMs a theoretical maturity and flexibility to rival the representational capacity of existing neural architectures. However, the HAM is still young and has yet to establish itself as a viable alternative to current Deep Learning architectures.

\subsection{Other models for Associative Memory} \label{sec:other-architectures}

The term ``associative memory'' has become a catch-all term for many different types of Content Addressable Memories (CAMs), including models like Sparse Distributed Memories~\cite{jaeckel1989alternative,kanerva1988sparse}, Memory Networks~\cite{weston2015Memory}, Key-Value Memories~\cite{tyulmankov2021Biological,miller2016KeyValue}, 
and Boltzmann Machines~\cite{hinton1984boltzmann,ackley1985learning,hinton2002Training}. Even the attention mechanism in Transformers~\cite{vaswani2017attention} is a differentiable form of associative memory~\cite{ramsauer2022Hopfield} where tokens act as queries to retrieve values stored by the other tokens. 
To be considered in this survey, a CAM must be a fixed-point attractor with tractable energy, like the paradigmatic Hopfield Net.
\begin{table*}[ht]
    \def\arraystretch{1.5}
    \setlength{\tabcolsep}{18pt}
    \centering

    \caption{Summarizing the similarities and differences between Diffusion Models and Associative Memory. Fields marked with a $^*$ indicate caveats. See \autoref{sec:differences} for details.}
    \begin{tabular}{l c c}
    \toprule
         & \textbf{Diffusion} & \textbf{Associative Memory} \\
         \midrule
      Parameterizes the$\ldots$ & Score function $\vf_{\vtheta}$ & Energy function $E_{\vtheta}$ \\
      Continuous Update  & $\tau \frac{d\vx}{dt} = \vf_{\vtheta}(\vx)$ & $\tau \frac{d\vx}{dt} = -\nabla_{\hat{\vx}} E_{\vtheta}(\hat{\vx})$ \\
      Discrete Update   & $\vx^{t+1} = \vx^t + \alpha \vf_{\vtheta}(\vx^t)$ & $\vx^{t+1} = \vx^t - \alpha \nabla_{\hat{\vx}} E_{\vtheta}(\hat{\vx}^t)$ \\
      Valid Time Domain & $t \in [0,T]$ & $t \geq 0$ \\
      Fixed Point Attractor? & No$^*$ & Yes\\
      Tractable Energy? & No$^*$ & Yes \\
      Undoes Corruption of $\ldots$ & Noise it was trained on$^*$ & Any kind \\
      \bottomrule
    \end{tabular}
    \label{tab:sim-and-diffs}
\end{table*}

\section{The Uncanny Resemblance of AMs and DMs}
\label{sec:similarities}

Associative Memories are different from Diffusion Models in that they are not primarily understood as generative models. 
However, just like the reverse process of DMs, memory retrieval can be easily be construed as a ``generation process'' that samples from some learned distribution. This realization makes it possible to directly compare AMs to DMs.

In this survey, we have emphasized the general definition of Associative Memories as \textit{any} parameterized Lyapunov energy: by only computing energy gradients, DMs are fundamentally different from AMs because they have no Lyapunov function guaranteeing stable convergence. Thus, DMs require tricks that alter dynamic process to ensure convergence (we tabulate the differences in more detail in \autoref{tab:sim-and-diffs}). Yet beyond this both methods share incredible similarities: 

\begin{itemize}[leftmargin=*]
    \item \textbf{Both model the energy.} DMs learn a parameterized score function $\vf_{\vtheta}$ to approximate the gradient of some true energy function $E$ such that $\vf_{\vtheta}(\vx) \approx -\nabla_\vx E(\vx) \; \forall \vx$. In AMs, this energy function is explicit and is defined by using architectures that directly model the energy $E_{\vtheta}$.
    
    \item \textbf{Both generate samples by descending the predicted gradient of the energy.} DMs directly output the estimated score $\vf_{\vtheta} \approx -\nabla_\vx E(\vx)$, whereas AMs will directly output a smooth energy $E_{\vtheta}(\hat{\vx})$ on which the gradient $-\nabla_{\hat{\vx}} E_{\vtheta} (\hat{\vx})$ can be directly calculated and descended. The update rules of both are identical as in \autoref{eq:continuous-diffusion-energy} and \autoref{eq:continuous-am-dynamics}.
    
    \item \textbf{Both converge to a solution that lies in the neighborhood of a local energy minimum.} In DMs, this behavior is a consequence of the manner in which it is trained: the final output $\vx^T$ exists in a region such that a small perturbation in any direction would increase its energy. In AMs, this statement is a requirement of the Lyapunov function; if the dynamics run for a sufficiently long time, we are guaranteed to reach a fixed point $\vx^\star$ that lies at a local energy minimum.
\end{itemize}

\subsection{Situations of Precise Equivalence} \label{sec:math-connection}

Concurrently to this work, \cite{ambrogioni2023search} has uncovered a precise mathematical connection between DMs and the Modern Hopfield Network (MHN). Specifically, if we assume $N$ discrete stored patterns and a time dependent temperature $\beta^{-1}_t = (T- t) \sigma_t^2$ that captures both the noise schedule $\sigma_t$ and a DM's constrained dynamics of $t < T$, an energy function that produces DM dynamics~\cite{raya2024spontaneous} is mathematically equivalent (up to a constant) to a MHN with exponential energy and time dependence:

\begin{equation}
    \label{eq:diffusion-energy-is-mhn}
E^\mathrm{DM}(\mathbf{x}, t) = -\sigma_t^2 \log \left( \frac{1}{N} \sum\limits_{n=1}^N e^{-\frac{||\mathbf{x} - \mathbf{y}_n ||^2_2}{2(T-t)\sigma_t^2}} \right) \iff -\beta^{-1}_t\log \left( \sum\limits_{n=1}^N e^{\beta_t \mathbf{x}^\intercal \mathbf{y}_n} \right) + \frac{||\mathbf{x}||_2^2}{2} = E^\mathrm{MHN}(\mathbf{x}, t).
\end{equation}

\noindent From this equation, \cite{ambrogioni2023search} claims that DM theory ``can be seen as a wide generalization of the classical theory of associative memory''. In this work, we emphasize a different viewpoint: DMs can be seen as a particular extension of AM theory that bypasses the stability guarantees of a Lyapunov energy by intelligently scheduling the noise (temperature) of a generative process that performs memory retrieval. In the single layer MHN, it is possible to find an energy function that will approximate it (e.g., the mixture-of-Gaussian like energy of \autoref{eq:diffusion-energy-is-mhn}); we believe that there is need to explore the mathematical equivalence of deeper DM architectures (e.g., U-Nets~\cite{ronneberger2015UNet}) and hierarchical AMs~\cite{krotov2021Hierarchical}.

\subsection{Reconciling the Differences} \label{sec:differences}

DMs and AMs are certainly not equivalent methods. 
However, we discover evidence that the theoretical gaps between DMs and AMs are not so significant in practice. 

\begin{itemize}[leftmargin=0cm]
    \item[] \textbf{Observation 1: DMs approximate fixed point attractors.} 
        Though the dynamics of DMs have no theoretical guarantees of fixed point attractors, we notice that the design of DMs seems to intelligently engineer the behavior of fixed-point attractors without constraining the architecture itself. We identify two fundamental tricks used by DMs that help approximate stable dynamics:

        \begin{description}
            \item[Trick 1] DMs explicitly halt their reconstruction process at time $t=T$ (i.e., requiring $\vx^\star = \vx^T$ and added noise $\veta(T)=0$) and are thus only defined for time $t \in [0, T]$. 
            $\vx^T$ then represents a \textit{contrived fixed point} because no further operations change it. We can say that $\vx^{t\neq T}$ corresponds to a data point with some corruption and $\vx^T$ corresponds to a \textit{memory} in the language of AMs.

            \item[Trick 2] We know that $\vx^T$ approximates a local energy minimum because of the \textit{noise annealing} trick introduced by~\cite{song2019Generative} and used by all subsequent DMs. During training, datapoints are perturbed with gradually increasing amounts of noise such that small noise is added around the point itself. This leads to a robust approximation of the true energy gradient localized around each data point where the original data point lies at the minimum. \cite{ambrogioni2023search} proved that this technique is equivalent to temperature annealing a softmax activation function during memory retrieval in the MHN.
        \end{description}

        \;\;\;We additionally see evidence of DMs storing ``memories'' that are actual training points. \cite{carlini2023Extracting} showed that one can retrieve training data almost exactly from publicly available DMs by descending an energy conditioned on prompts from the training dataset. It seems that this behavior is particularly evident for images considered outliers and for images that appear many times. Viewing DMs as AMs, this behavior is not surprising, as the whole function of AMs is to retrieve close approximations to data it has seen before.

        \;\;\;Tricks 1 \& 2 imply that DMs are inescapably bound to a knowledge of the current time $t$. The time $t$ defines both the total noise the model should expect in a given signal (i.e., the standard deviation or the ``width'' of Gaussian peaks around each data point) and how much noise the model should expect to remove (i.e., the step size down the energy). Given a signal with an unknown quantity of noise, a user must either make a ``best guess'' for the time $t$ corresponding to this amount of noise, or restart the dynamics at time $t=0$ to reset the ``noise scheduler'', which causes the model to make large jumps around the energy landscape and likely land it in a distant energy minimum.
        Currently, AMs have no such dependence between corruption levels and time $t$, though it is easy to include time-dependence into the memory retrieval dynamics of AMs.

    \vspace{0.2cm}
    \item[] \textbf{Observation 2: DMs can undo more than Gaussian noise.}
        In order for a DM to behave like an AM, it must be possible to undo any kind of corruption (e.g., inpainting, blur, pixelation, etc.), not just the white- or Gaussian-noise associated with Brownian motion as originally formulated in \cite{sohl-dickstein2015Deep,song2019Generative};
        this is because all corruptions are a form of error that causes samples to have higher energy.
        \cite{voleti2022Scorebased,nachmani2021Non} showed that the performance of DMs can actually improve when considering other types of noisy corruption in the forward process. However, it also seems that DMs can learn to reverse any kind of corrupting process. 
        \cite{bansal2022Cold} demonstrates that DMs can be trained to invert arbitrary image corruptions that generate samples almost as well as those trained to invert only Gaussian noise introduced by the Brownian motion of the forward process presented in this work. Thus, DMs exhibit AM behavior by following a general score function that can seemingly recover ``fixed points'' of some energy landscape by undoing arbitrary corruptions.

    \vspace{0.2cm}
    \item[] \textbf{Observation 3: Unconstrained DMs work best with special architectures.}
        One advantage of DMs over AMs is that they are ``unconstrained'' and can use any neural network architecture to approximate the score function; that is, the architecture is not required to be the gradient of an actual scalar energy. The only requirement is that the neural network chosen to approximate the score must be \textit{isomorphic} such that the function's output is the same shape as its input (i.e., $\vf_{\vtheta}: \R^d \mapsto \R^d$). However, not all isomorphic architectures are created equal and only select architectures are used for DMs in practice. Both standard feedforward networks~\cite{salimans2021Should} and vanilla Transformers have struggled to generate quality samples using diffusion~\cite{bao2022All,yang2022Your}.
        Most applications of DMs use some modification of the U-Net architecture~\cite{ronneberger2015UNet} originally used by \cite{ho2020Denoising}, though the original DM paper \cite{sohl-dickstein2015Deep} used shallow MLPs, and recent work \cite{bao2022All} has shown how to engineer vision Transformers~\cite{dosovitskiy2020Image} to achieve a similar reconstruction quality as U-Nets on images.

    \vspace{0.2cm}
    \item[] \textbf{Observation 4: DMs work with explicit energy.}
        Though DMs characterize an energy landscape by modeling its gradient everywhere, they do not inherently have a concept of the energy value itself. However, \cite{song2020ScoreBased} showed that one can actually compute an exact energy for DMs using the instantaneous change of variables formula from \cite{chen2018Neural}, with the caveat that this equation is expensive to compute. Estimations of the energy are preferred over direct computation in practice~\cite{grathwohl2018FFJORD}.

        \;\;\;Another approach for enforcing an energy on DMs is to choose an architecture that parameterizes an actual energy function, whose score function then describes a conservative gradient field. \cite{salimans2021Should} researched exactly this, exploring whether a generic learnable function $\vff_{\vtheta}(\vx;t)$ that is constrained to be the true gradient of a parameterized energy function as in \autoref{eq:constrained-ebms} is able to attain sample quality results similar to those of unconstrained networks.

        \begin{equation}
            \label{eq:constrained-ebms}
            E_{\vtheta}(\vx; t) = \frac{1}{2\sigma(t)} || \vx - \vff_{\vtheta}(\vx;t) || ^2
        \end{equation}

        The score $\vf_{\vtheta}$ of this energy can be written by computing the analytical gradient

        \begin{equation}
        \label{eq:score-of-constrained-ebms}
        \vf_{\vtheta}(\vx; t) = \frac{\vx - \vff_{\vtheta}(\vx; t)}{\sigma(t)}  \nabla_\vx \vff_{\vtheta}(\vx;t) \\- \frac{\vx - \vff_{\vtheta}(\vx; t)}{\sigma(t)} \,.
        \end{equation}

        \cite{salimans2021Should} notes that the second term of the equation is the standard DM, while the first term involving $\nabla_\vx \vff_{\vtheta}(\vx;t)$ is new and helps guarantee that $\vf_{\vtheta}(\vx; t)$ is a conservative vector field.  They showed that constraining the score function to be the gradient of the energy in \autoref{eq:constrained-ebms} does not hurt generation performance and provides hope that AMs with constrained energy can one day match the performance of unconstrained DMs.

\end{itemize}

\section{Conclusions \& Open Challenges}
\label{sec:conclusions}

Diffusion Models and Associative Memories have remarkable similarities when presented using a unified mathematical notation: both aim to minimize some energy by following its gradient. The output of both approaches represents some sort of \textit{memory} that lies in a local minimum (maximum) of the energy (log-probability). However, these approaches are not identical as evidenced by different validity constraints on architectures and time domains. The training philosophy behind each approach is also different: DMs assume that the energy function is intractable and learn the gradient using known perturbations of data as the objective, while AMs focus on learning the fixed points of a tractable energy.

\subsection{Directions for Associative Memories}
AMs have not gained the traction of DMs in AI applications. Many researchers in the field focus on the  {\tinybrain} architecture, trying to improve its theoretical memory capacity~\cite{millidge2022Universal,burns2023Simplicial} or apply related models to modern problems~\cite{liang2021Can}. Other researchers are connecting the memory-retrieval capabilities of AMs with existing feed-forward architectures like the Transformer~\cite{ramsauer2022Hopfield,widrich2020Modern}; in doing so they discard the idea of global optimization on the energy. 
In part, these other research directions exist because no pure AM has shown impressive performance on large data until \cite{hoover2023Energy} introduced the ``Energy Transformer'' that is a true Lyapunov energy (where the forward-pass through the network is gradient descent down this energy) that can be trained and used in the same manner as conventional Transformers. However, even this AM does not yet show significant performance gain over traditional methods.

The empirical success of DMs across many domains should provide hope that modern AM architectures~\cite{krotov2021Hierarchical,hoover2022Universal} can achieve performance parity on similar tasks. Constrained DMs show no worse performance than unconstrained models~\cite{salimans2021Should}, and HAM theory \cite{krotov2021Hierarchical} can build AMs that resemble U-Nets~\cite{hoover2022Universal} and Transformers~\cite{hoover2023Energy}. 

\subsection{Directions for Diffusion Models}
DM researchers should find the theoretical framework of the Lyapunov function from AMs compelling: in the absence of the Lyapunov function, DMs must manufacture fixed points using dynamic noise schedules and step sizes. 
However, if the score function of DMs is shown to be a conservative vector field as in~\cite{salimans2021Should}, perhaps DMs have learned fixed point structure and can behave well in all time $t>T$. Viewing DMs as fixed-point attractors like AMs would additionally allow us to theoretically characterize its memory capacity (where ``memory capacity'' can be seen as a proxy for the ``scaling laws'', see \autoref{sec:scaling-laws}). Finally, viewing the sampling procedure from DMs as a form of gradient descent (as in AMs) allows optimization methods like ADAM~\cite{kingma2017Adam} and L-BFGS~\cite{liu1989limited} to be used for to sample from the probability distribution.

\subsection{Scaling Laws from the Perspective of Associative Memory}
\label{sec:scaling-laws}

The performance of Transformers on language is famously characterized by the ``scaling laws'', which claim that a model's performance will improve as a power-law with model size, dataset size, and the amount of compute used for training~\cite{kaplan2020Scaling}. DMs exhibit similar scaling behaviors~\cite{nichol2021Improved}, where larger models perform better than smaller ones in a predictably improving trend. However, the ``scaling laws'' are empirical only, and there is no agreed-upon theory to justify why a model's performance would continue to grow with the model size. 

AMs offer one possible answer by characterizing large-model performance as one of \textit{memory capacity} (see \autoref{sec:memory-capacity}), as recently noted by \cite{cabannes2023Scaling,niu2024Scaling}. In the world of AMs, each parameter can be seen as an ``attractor'' in the data space; thus, more parameters means more local energy minima (memories). Similarly, more data means the parameters can identify more meaningful local minima in the energy. More compute, as measured in terms of model depth or number of iterations down an energy landscape, means more optimal (lower energy) retrievals. These hypotheses are still unexplored research questions that come from intuitively understanding large models as AMs.

\subsection{Closing Remarks}

Very few researchers will observe the rapid advances of AI today and notice a trend towards the dynamical processes of Associative Memories first established by John Hopfield in the 1980s. However, many of the theoretical guarantees of Associative Memories are captured in the design of increasingly popular Diffusion Models that have proven themselves fixtures for many applications of generative modeling.
This survey is a first step towards a more comprehensive understanding of the connections between Diffusion Models and general Associative Memories. We hope that our work inspires further research into these exciting fields and that it helps to foster a new generation of AI systems that are capable of unlocking the secrets of memory and perception. 
\bibliographystyle{unsrt}
\bibliography{bibliography}

\end{document}